\documentclass{article}

\usepackage[margin=1in]{geometry}

\usepackage{graphicx} % more modern
\usepackage{subfigure} 

\usepackage{natbib}

\usepackage{algorithm}
\usepackage{algorithmic}

\usepackage{hyperref}

\usepackage[T1]{fontenc}
\usepackage{graphics}
\usepackage{amsmath, amssymb,amsthm}
\usepackage{tikz}
\usetikzlibrary{positioning}
\newtheorem{lemma}{Lemma}
\newtheorem*{lemmaNoNum}{Lemma}
\newtheorem{corollary}{Corollary}
\newtheorem{definition}{Definition}

\newtheorem{theorem}{Theorem}
\definecolor{deanPURPLE}{rgb}{.3,0,.5}
\definecolor{jrBLUE}{rgb}{.2, .2, .8}
\definecolor{gray}{rgb}{.8,.8,.8}

\newcommand{\diag}{\hbox{diag}}

\begin{document}

\title{Spectral dimensionality reduction for HMMs}
\author{Dean P. Foster\\University of Pennsylvania \and Jordan Rodu\\University of Pennsylvania \and Lyle H. Ungar\\University of Pennsylvania}
\vskip 0.3in

\tikzstyle{state}=[shape=circle,draw=blue!50,fill=blue!20]
\tikzstyle{observation}=[shape=circle,draw=blue!50,fill=red!20,thick]
\tikzstyle{lightedge}=[<-,thick,red]
\tikzstyle{mainstate}=[state,thick]
\tikzstyle{mainedge}=[<-,thick]

\maketitle
\begin{abstract}

Hidden Markov Models (HMMs) can be accurately approximated using
 co-occurrence frequencies of pairs and triples of observations by
 using a fast spectral method \cite{hsu} in contrast to the usual slow
 methods like EM or Gibbs sampling.  We provide a new spectral method
 which significantly reduces the number of model parameters that need
 to be estimated, and generates a sample complexity that does not
 depend on the size of the observation vocabulary.  We present an
 elementary proof giving bounds on the {\em relative} accuracy of
 probability estimates from our model.  (Correlaries show our bounds
 can be weakened to provide either L1 bounds or KL bounds which
 provide easier direct comparisons to previous work.)  Our theorem
 uses conditions that are checkable from the data, instead of putting
 conditions on the unobservable Markov transition matrix.

\end{abstract}

\section{Introduction}

For many applications such as language modeling, it is useful to
estimate Hidden Markov Models (HMMs) \cite{rabiner1989tutorial} in which observations drawn from
a large vocabulary are generated from a much smaller hidden state.
Standard HMM estimation techniques such as Gibbs sampling \cite{geman} and
EM \cite{baum1970maximization,dempster1977maximum} methods, although
very  widely used, can require some effort to apply as 
they are often either slow or prone to get stuck in local optima. 
Hsu, Kakade and Zhang, in a path breaking paper, \cite{hsu} showed that
HMMs can, in theory, be efficiently and 
accurately estimated using closed form calculations on trigrams of
observations which have been projected onto a low dimensional space.
Key to this approach is the use of singular value decomposition (SVD)
on the matrix of covariances between adjacent observations to learn a
matrix $U$ that projects observations onto a space of the same dimension 
as the hidden state. 
Perhaps surprisingly, co-occurrence statistics on
unigrams, pairs, and triples of observations are sufficient to
accurately estimate a model equivalent to the original HMM.

The true hidden state itself cannot, of course, be estimated (it is not observed),
but one can estimate a linear transformation of the hidden state  which contains
sufficient information to give an optimal (in a sense to be made precise
below) estimate of the probability of any sequence of
observations being generated by the HMM~\cite{hsu}. 
The method of \cite{hsu}, and the extensions to it presented in this paper 
do not require any EM or Gibbs
sampling, but only need an SVD on bigram observation counts. 
Since SVD is an efficient method guaranteed to
return the correct result in a known number of steps, this is a major advantage
over the iterative EM method.

Hsu et al. \cite{hsu} estimate a size $mv$ matrix mapping between the
the dimension $v$ observation space and a reduced dimension space of 
size $m$ (the dimension of the hidden state space).
They also need to
estimate a tensor of size $vm^2$. We provide an alternate formulation
that replaces their $vm^2$ tensor with one of size $m^3$. Since
the observation vocabulary, $v$, is often much larger than the
state space ($v \gg m$), this provides significant reduction in model size, and hence, as 
we show below, in sample complexity.

\subsection{HMM set-up and notation}

We now introduce the notation and model used throughout our paper.

Consider an HMM where $T$ is an $m \times m$ transition matrix
on the hidden state, $O$ is a $v \times m$ emission matrix giving
the probabilities of hidden state $h=j$ emitting observation $x=i$,
and $\pi$ is a vector of initial state probabilities in which  $\pi_i$ is
the probability that $h_1=i$. Jaeger \cite{jaeger} showed that the joint
probability of a sequence of observations from this HMM is given by
\begin{equation}
\label{eq:jaeger}
 Pr(x_1, x_2, \ldots, x_t) = 1^\top A_{x_t} A_{x_{t-1}} \cdots A_{x_1} \pi ,
\end{equation}
where  $A_{x} \equiv T \diag(O^\top \delta_{x})$,
 $\delta_x$ is the unit vector of length $v$  with a single 1 in the $x$th position
and $\diag(v)$ creates a matrix with the elements of the vector $v$ on its diagonal
and zeros everywhere else.

$A_t$ is called an 'observation operator', an idea dating back to multiplicity automata \cite{schutzenbeegeb1961definition,carlyle1971realizations,fliess1974matrices}, and foundational in the theory of Observable Operator Models \cite{jaeger} and Predictive State Representations \cite{littman2002predictive}. It is
effectively a third order tensor, giving the distribution vector over
states at time $t+1$ as a function of the state distribution vector at
the current time $t$ and the current observation $\delta_{x_t}$.
Since $A_t$ depends on the hidden state, it is not observable, 
and hence cannot be directly estimated.  But
\cite{hsu} showed that under certain conditions there exists a fully
observable representation of the observable operator model.  We now present a novel, fully reduced dimensional version of the observable representation.

%%%%%%%%%%%%%%%%%%%%%%%%%%%%%%%%%%%%%%%%%%%%%%%%%%%%%%
\subsection{The reduced dimension model}

Define a random variable
$y_t=U^\top\delta_{x_t}$,
%These $z$'s are low dimensional representations of $\delta_x$'s.  
where $U$ has orthonormal columns and is a matrix mapping from observations to the reduced
dimension space.

We show below that  
\begin{eqnarray}
Pr(x_1 ,x_2, \ldots , x_t) = c_\infty^\top C_{y_t} C_{y_{t-1}} \cdots C_{y_1} c_1 \label{eqn:reduced}
\end{eqnarray}
holds where
\begin{eqnarray*}
c_1 &=& \mu\\
c^\top_\infty &=&  \mu^\top \Sigma^{-1}\\
C_y\equiv C(y) &=& K(y)\Sigma^{-1}
\end{eqnarray*}
and $\mu$  = $E(y_1)$, $\Sigma$ =  $E(y_2 y_1^\top)$, and 
$K(a) = E(y_3 y_1^\top y_2^\top) a$ are easy to estimate using the method of moments.\footnote{
Note that $K()$ is a tensor. When multiplied by a vector $a$, it produces
a matrix. $K()$ is linear in each of the three reduced dimension observations,
$y_1$, $y_2$ and $y_3$.}

The matrix $U$ can be derived in several ways; \cite{hsu} show that taking it to consist of
the left singular vectors of $P_{21}$ corresponding to the largest
singular values gives good properties, where $P_{21}$ is a matrix such that $[P_{21}]_{ij} = Pr[x_2=i, x_1=j]$. The matrix $U$ and its properties will be discussed in more detail below.
 
Note that the model ($c_1, c_\infty, C(y)$) will be estimated using only
trigrams.  Once a model has been learned, the probability of any
observed sequence $(x_1, x_2, . . . x_t)$ can be computed using
equation \ref{eqn:reduced}, or the conditional probability $Pr(x_t|x_1,
x_2 . . . x_{t-1})$ of the next observation $x_t$ in a sequence can be
computed by $Pr(x_t|x_{1:t-1})= c_\infty^\top C(y_t)c_t$ with
recursive updates $c_{t+1} = C(y_t) c_t/(c^\top_\infty C(y_t) c_t)$.
The key term in the model is thus $C(y)$, which can be viewed as a
tensor which takes as input the current observation $x_t$ and produces
a matrix which maps (after normalization) from the current ``hidden state estimate'' $c_t$
to the next one $c_{t+1}$. More precisely, 
$c_{t+1} = (U^\top O) \widehat{h}_{t+1} (x_{1:t})$ is a linear function of the
conditional expectation of the unobservable hidden state  $\widehat{h}_{t+1}(x_{1:t})$, which is the conditional probability vector over states at time $t+1$. 

%%%%%%%%%%%%%%%%%%%%%%%%%%%%%%%%%%%%%%%%%%%%%%%%%%%%%%
\subsection{Comparison to Hsu et al.}

Hsu et al. \cite{hsu} derive a similar model which we state here for comparison.

%\begin{displaymath}
\begin{equation}
Pr(x_1 ,x_2, ... , x_t) = b_\infty^\top B_{x_t} B_{x_{t-1}} \ldots B_{x_1} b_1 \label{eqn:hsu}
\end{equation}
%\end{displaymath}
where
\begin{eqnarray*}
b_1 & = & U^\top P_1 \\
b^\top_\infty & = & P_1^\top (U^\top P_{21})^+\\  
B_x & = & (U^\top P_{3x1})(U^\top P_{21})^+   
\end{eqnarray*}
and $[P_1]_i = Pr[x_1 = i]$, $P_{21}$ as defined above, and 
$[P_{3x1}]_{ij} = Pr[x_3=i, x_2=x, x_1=j]$ are the frequencies of
unigrams, bigrams, and trigrams in the observed data. Note that the subscripts on $x$ refer to their
positions in trigrams of observations of the form $(x_1, x_2, x_3)$.

Our major modeling change will be to replace $B_x$ in equation \ref{eqn:hsu} with the 
lower dimensional tensor $C(y)$ which depends on the reduced dimension
projection $y \equiv U^\top \delta_{x}$ instead of the unreduced $x$. The models are easily related by the following lemma:

\begin{lemma}\label{lem:representation} Assume the hidden state is of dimension $m$ and the rank
of $O$ is also $m$. Then:
\begin{eqnarray}
Pr(x_1 ,x_2, \ldots, x_t)
      &=& 1^\top A_{x_t} A_{x_{t-1}} \cdots A_{x_1} \pi \label{eqn:orig}\\
      &=& b_\infty^\top B_{x_t} B_{x_{t-1}} \cdots B_{x_1} b_1 \label{eqn:hsu:lemma} \\
      &=& c_\infty^\top C_{y_t} C_{y_{t-1}} \cdots C_{y_1} c_1 \label{eqn:new}
\end{eqnarray}
Where (\ref{eqn:hsu:lemma}) requires $U^\top O$ to be invertible, and
(\ref{eqn:new}) requires $\hbox{range}(O) \subset \hbox{range}(U)$.\footnote{
If the matrix $U$ is formed from the left singular vectors of $P_{21}$
corresponding to nonzero singular values, then it will satisfy this condition;
See~\cite{hsu} lemma 2.}
\end{lemma}

{\bf Proof sketch:} 
 Paper \cite{jaeger} showed (\ref{eqn:orig}), paper \cite{hsu} showed
(\ref{eqn:hsu:lemma}), and (\ref{eqn:new}) follows from a telescoping product of
the following items:
\begin{eqnarray*}
c_1 &=&  U^\top O \; \pi \\
c^\top_\infty &=&{\bf 1}^\top \;  (U^\top O)^{-1}\\
C_y = C(y) & = &  U^\top  O \; A_x \; (U^\top O)^{-1}
\end{eqnarray*}
where $y = U^\top \delta_x$.  More details are given in the
supplemental material.
\hfill $\Box$

%%%%%%%%%%%%%%%%%%%%%%%%%%%%%%%%%%%%%%%%%%%%%%%%%
%
%              FIGURE
%%%%%%%%%%%%%%%%%%%%%%%%%%%%%%%%%%%

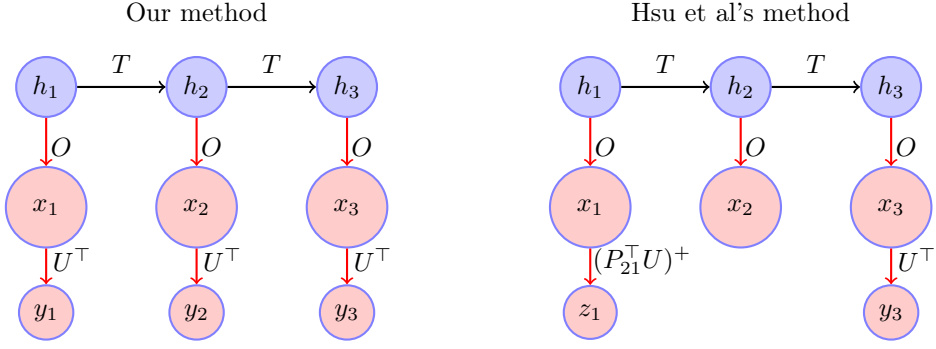
\begin{figure}[htbp]
\begin{center}
\hspace{-.5in}
\begin{tikzpicture}
% 0th column
\node (h0_1)        at (0.3,5) {};
% 1st column
\node[mainstate] (h1_1) at (2,5) {$h_1$}
;%	edge[mainedge](h0_1);  % adds an edge into the figure
\node		   at (2.2,4.2) {$O$};
\node [observation] (x1_1)  at (2,3.4) {$\;\; x_1 \;\;$}
    edge[lightedge] (h1_1);
\node		   at (2.2,2.7) {$\;\;\;U^\top$};
\node [observation] at (2,2) {$y_1$}
    edge[lightedge] (x1_1);

%% half column
\node		   at (3,5.3) {$T$};
% 2nd column
\node               at (4,6) {Our method};
\node[mainstate] (h1_2) at (4,5) {$h_2$}
    edge[mainedge] (h1_1);
\node		   at (4.2,4.2) {$O$};
\node [observation] (x2_2)  at (4,3.4) {$\;\; x_2 \;\;$}
    edge[lightedge] (h1_2);
\node		   at (4.2,2.7) {$\;\;\;U^\top$};
\node [observation] at (4,2) {$y_2$}
    edge[lightedge] (x2_2);
\node		   at (5,5.3) {$T$};
% 3rd column
\node[mainstate] (h1_3) at (6,5) {$h_3$}
    edge[mainedge]  (h1_2);
\node		   at (6.2,4.2) {$O$};
\node [observation] (x3_3)  at (6,3.4) {$\;\; x_3 \;\;$}
    edge[lightedge] (h1_3);
\node		   at (6.2,2.7) {$\;\;\;U^\top$};
\node [observation] at (6,2) {$y_3$}
    edge[lightedge] (x3_3);
% and beyond
\node              at (7.7,5) {}
;%    edge[mainedge] (h1_3);  % removing edge leaving graph
\end{tikzpicture}
\hspace{-.25in}
\begin{tikzpicture}
% 0th column
\node (h0_1)        at (0.3,5) {};
% 1st column
\node[mainstate] (h1_1) at (2,5) {$h_1$}
;%	edge[mainedge](h0_1);  % adds an edge into the figure
\node		   at (2.2,4.2) {$O$};
\node [observation] (x1_1)  at (2,3.4) {$\;\; x_1 \;\;$}
    edge[lightedge] (h1_1);
\node		   at (2.2,2.7) {$\;\;\;\;\;\;\;\;\;\;(P_{21}^\top U)^+$};
\node [observation] at (2,2) {$z_1$}
    edge[lightedge] (x1_1);
%% half column
\node		   at (3,5.3) {$T$};
% 2nd column
\node               at (4,6) {Hsu et al's method};
\node[mainstate] (h1_2) at (4,5) {$h_2$}
    edge[mainedge] (h1_1);
\node		   at (4.2,4.2) {$O$};
\node [observation] (x2_2)  at (4,3.4) {$\;\; x_2 \;\;$}
    edge[lightedge] (h1_2);
\node		   at (5,5.3) {$T$};
% 3rd column
\node[mainstate] (h1_3) at (6,5) {$h_3$}
    edge[mainedge]  (h1_2);
\node		   at (6.2,4.2) {$O$};
\node [observation] (x3_3)  at (6,3.4) {$\;\; x_3 \;\;$}
    edge[lightedge] (h1_3);
\node		   at (6.2,2.7) {$\;\;\;U^\top$};
\node [observation] at (6,2) {$y_3$}
    edge[lightedge] (x3_3);
% and beyond
\node              at (7.7,5) {}
;
\end{tikzpicture}
\end{center}
\caption{ Two HMMs with states $h_1$, $h_2$, and $h_3$ which emit
observations $x_1$, $x_2$, and $x_3$.  On the left, they are further projected onto lower dimensional
space with observations $y_1$, $y_2$, $y_3$ by $U$ from which our core
statistic $C_y$ is computed based on $K=E(y_3y_1^\top y_2^\top)$ which
is a $(m\times m\times m)$ tensor.  On the right, $x_1$ is hit by
$(P_{21}^\top U)^+$ to make a lower dimensional $z_1$, $x_2$ is left
unchanged and $x_3$ has its dimension reduced by $U^{\top}$.  These
terminal leafs are then used by \cite{hsu} to estimate their $B_x$ via
estimating $E(y_3 z_1^\top \delta_{x_2}^\top)$ which is a tensor of
size $(m\times m\times v)$.}
\end{figure}
%%%%%%%%%%%%%%%%%%%%%%%%%%%%%%%%%%%%%%%%%%%%%%%%%%%%%%%%%%%%%%%%%%%%%%%%%%%

We improve \cite{hsu} in three ways:
\begin{enumerate}
\item By reducing the size of the matrix that is estimated, we can
 achieve a lower sample complexity.  In particular, our sample
 complexity does not depend on the size of the vocabulary nor on the
 frequency distribution of the vocabulary.
\item Since the conditions given in \cite{hsu} are in terms of the
transition matrix $T$, they can not be checked.  We  instead focus on
conditions that are checkable from the data. 
\item Instead of using either a L1 error or a relative entropy error,
 we estimate the probabilities with relative accuracy.  In other
 words,  we show that $|\widehat{p} - p|/p$ is smaller than $\epsilon$.  This often is a
 more useful bound than just knowing $|\widehat{p} - p|$ is small.  For
 example, it implies that computing conditional probabilities are
 off by less than $2 \epsilon$.  Both L1 and relative entropy errors
can be computed from these bounds.
\end{enumerate}

Our main theorem is weaker (as stated) than \cite{hsu} in that we
 assume knowledge of $U$ rather than estimating it from a thin SVD of
 $P_{21}$ as they do.  Since the accuracy lost when estimating $U$ is
 identical to that given in their paper, we will not discuss it here.

%%%%%%%%%%%%%%%%%%%%%%%%%%%%%%%%%%%%%%%%%%%%%%%%%%%%%%%%%%%%%%%%%%

\section{Theorems}

The remainder of this paper presents one main theorem giving finite sample
bounds for our reduced dimensional HMM estimation method.  We first derive
these in terms of properties of the first three moments of the reduced rank $Y$'s, 
where $Y$ is the random variable which takes on values of the reduced rank 
observation $y = U^\top \delta_x$. We then convert those bounds to be in terms
of the estimates, rather than the unobservable true values, of the model.

%% Due to space limitations, some of the proofs are contained in supplementary
%% materials.

Our general strategy of estimating $\Pr(x_t, x_{t-1},\ldots, x_1)$ is
via the method of moments.  We have $\Pr()$ written in terms of $c_\infty^\top$,
$c_1$ and $C(y_t)$.  Since each of these three items can be written in
terms of moments of the $Y$'s we can plug in these moments to generate
an estimate of $\Pr()$.  Thus we can define:
\begin{equation}
\label{eq:phat}
\widehat{\Pr}(x_t, x_{t-1},\ldots, x_1) = \widehat{c}^\top_\infty
\widehat{C}(y_t)\widehat{C}(y_{t-1})\cdots\widehat{C}(y_1) \widehat{c}_1 
\end{equation}

where
\begin{eqnarray*}
\widehat{c}_1 & = & \widehat{\mu} \\
\widehat{c}^\top_\infty & = &  \widehat{\mu}^\top \widehat{\Sigma}^{-1}\\
\widehat{C}(y) & = & \widehat{K}(y) \widehat{\Sigma}^{-1}
\end{eqnarray*}
where $\widehat{\mu}$, $\widehat{\Sigma}$ and $\widehat{K}()$ are the
empirical estimates of the first, second and third moments of the
$Y$'s, namely $\widehat{\mu}= \frac{1}{N}\sum_{i=1}^N Y_1^{(i)}$, $\widehat{\Sigma}=\frac{1}{N}
\sum_{i=1}^N Y_1^{(i)}Y_2^{(i)\top}$, $\widehat{K}(y) = \frac{1}{N}\sum_{i=1}^N
Y_1^{(i)}Y_3^{(i)\top} Y_2^{(i)\top} y$, where $Y^{(i)}$ indexes the
$N$ different independent observations of our data. 
These moments estimate the mean vector $\mu$, the variance matrix $\Sigma$, and the skewness tensor $K()$.

\begin{definition} Define $\Lambda$ as the smallest element of
$\mu$, $\Sigma^{-1}$ and $K()$.  In other words,
\begin{displaymath}
\Lambda \equiv \min\{\min_i |\mu_i|,\min_{i,j} |\Sigma_{ij}^{-1}|,
\min_{i,j,k} |K_{ijk}|\}
\end{displaymath}
where we define $K_{ijk} = K(\delta_j)_{ik}$ are the elements of the
tensor $K()$. Likewise we define the empirical version as
\begin{displaymath}
\widehat{\Lambda} \equiv \min\{\min_i |\widehat{\mu}_i|,\min_{i,j} |\widehat{\Sigma}_{ij}^{-1}|,
\min_{i,j,k} |\widehat{K}_{ijk}|\}
\end{displaymath}
\end{definition}
\begin{definition} Define $\sigma_m$ as the smallest singular value of
 $\Sigma$, and $\widehat{\sigma}_m$ the smallest singular value of
$\widehat{\Sigma}$. 
\end{definition}

The parameters $\Lambda$ and $\sigma_m$ will be central to our
analysis.  Theorem 1 gives sample complexity bounds on relative error
in estimating the probability of a sequence being generated from an
HMM as a function of $\Lambda$ and $\sigma_m$, and the following lemmas
reformulate those bounds into a more useful form in terms of their
estimates.  As quantified and proved below, both $\Lambda$ and
$\sigma_m$ must be ``sufficiently large''; when they approach zero
one loses the ability to accurately estimate the model.

If $\sigma_m = 0$ then $U^\top O$ will not be invertible, and one cannot
infer the full information content of the hidden state from its associated
observation, violating the condition required in \cite{hsu} for
(\ref{eqn:hsu:lemma}) to hold.  As $\sigma_m$ becomes increasingly close to
zero, it becomes increasingly hard to identify the hidden state, and more
observations are required. 
Problems with small $\sigma_m$ are intrinsically difficult.  As has
been pointed out by \cite{hsu}, some problems of estimating HMM's are
equivalent to the parity problem \cite{terwijn2002learnability}.  So for such data,
our algorithm need not perform well.  For parity-like problems,
$\sigma_m$ is in fact zero, or close to it; Hence we end up with a
useless bound for such hard problems.

If $\Lambda$ is close to zero, then even if the absolute error is
small, the relative error can be arbitrarily large, as it involves
dividing by the small true value of the parameter being
estimated. Fortunately, as discussed below, since $\Lambda$ depends on the
somewhat arbitrary matrix $U$, one can shift $\Lambda$ away from zero
by rotating and rescaling $U$.  

The proof of Theorem 1 is based on the idea that if we can
estimate each term in $\mu$, $\Sigma$ and $K()$ accurately on an
absolute scale (which will follow from basic central limit like
theorems) then we can estimate them on a relative scale if $\Lambda$
is large.  Hence, our main condition is that $\Lambda$ is bounded away
from zero.  In fact, if we take the usual statistical limit of having
the sample size $N$ go to infinity and holding everything else constant, then:
\begin{displaymath}
\left|\frac{\widehat \Pr (x_1, \ldots, x_t)}{\Pr (x_1, \ldots, x_t)}-1\right|
\leq   \frac{18 m t}{\sigma_m^2 \Lambda\sqrt{N}}\;\sqrt{\log (m/\delta)} 
\end{displaymath}
%\comment{Some algebra:
%\begin{eqnarray*}
%\Lambda & \geq & \frac{3 m}{\sigma_m^2(1 - \sqrt[2T+3]{1 +
%\epsilon})}\sqrt{\frac{8 \log \frac{m}{\delta}}{N}} \approx \\
%&& \approx  \frac{3 m}{\sigma_m^2 \epsilon/2T}\sqrt{\frac{8 \log \frac{m}{\delta}}{N}} \\
%\epsilon &\geq \approx&  \frac{6 m T}{\sigma_m^2 \Lambda}\sqrt{\frac{8 \log \frac{m}{\delta}}{N}} \\
%\epsilon &\geq \approx&  \frac{18 m T}{\sigma_m^2 \Lambda\sqrt{N}}\sqrt{\log (m/\delta)} \\
%\end{eqnarray*}}
with probability greater than $1 - \delta$ when $N$ is large enough.

The following theorem gives the finite sample bound in terms of a
sample complexity:      

\begin{theorem} \label{thm:main} Let $X_t$ be generated by an $m \ge 2$ state HMM.
Suppose we are given a $U$ which has the property that
$\hbox{range}( O) \subset \hbox{range}(U)$ and $|U_{ij}| \le 1$. Suppose we use
equation (\ref{eq:phat}) to estimate the probability based on $N$
independent triples.  Then
\begin{equation}
N \geq \frac{128 m^2 }{(\sqrt[2t+3]{1 +
\epsilon}-1)^2 \; \Lambda^2 \sigma_m^4}
\log \left(\frac{2m}{\delta}\right) 
\label{eq:sample:complexity}
\end{equation}
implies that
\begin{eqnarray*}
1-\epsilon\leq\left|\frac{\widehat \Pr (x_1, \ldots, x_t)}{\Pr (x_1, \ldots, x_t)}\right|\leq 1+\epsilon
\end{eqnarray*}
holds with probability at least $1 - \delta$.
\end{theorem}

Before proceeding with the proof of this theorem, we present and prove two corollaries that correspond directly to Theorems 6 and 7 of \cite{hsu}.

\begin{corollary}\label{cor:hsu6}
Assume Theorem \ref{thm:main} holds, then with probability at least $1-\delta$, 
\begin{eqnarray*}
\sum_{x_1, \ldots, x_t} |\widehat \Pr (x_t, \ldots, x_t) - \Pr (x_1, \ldots, x_t)|\leq \epsilon 
\end{eqnarray*} 
\end{corollary}

{\bf Proof of Corollary \ref{cor:hsu6}:} We have
\begin{align*}
&1-\epsilon\leq\left|\frac{\widehat \Pr (x_1, \ldots, x_t)}{\Pr (x_1, \ldots, x_t)}\right|\leq 1+\epsilon\\
\Rightarrow &\left|\frac{\widehat \Pr (x_1, \ldots, x_t)}{\Pr (x_1, \ldots, x_t)} -1 \right|\leq \epsilon\\
\Rightarrow &\left|\widehat \Pr (x_1, \ldots, x_t) - \Pr (x_1, \ldots, x_t) \right|\leq \epsilon\Pr (x_1, \ldots, x_t)\\
\Rightarrow &\sum_{x_1, \ldots, x_t}\left|\widehat \Pr (x_1, \ldots, x_t) - \Pr (x_1, \ldots, x_t) \right|\\
&~~~~~~~~~~~~~~~~\leq \epsilon\sum_{x_1, \ldots, x_t}\Pr (x_1, \ldots, x_t)\\
\Rightarrow &\sum_{x_1, \ldots, x_t}\left|\widehat \Pr (x_1, \ldots, x_t) - \Pr (x_1, \ldots, x_t) \right|\leq \epsilon\\
\end{align*}
\hfill $\Box$

\begin{corollary}\label{cor:hsu7}
Assume Theorem \ref{thm:main} holds, then we have
\begin{align*}
KL(\Pr(x_t|x_1, & \ldots x_{t-1})||\widehat \Pr (x_t|x_1, \ldots x_{t-1}))\\
&=E \left(\ln \frac{\Pr (x_t|x_1, \ldots x_{t-1})}{\widehat \Pr (x_t|x_1, \ldots x_{t-1})}\right)\leq 6\epsilon
\end{align*}
\end{corollary}

{\bf Proof of Corollary \ref{cor:hsu7}:} We have
\begin{align*}
&1-\epsilon\leq\left|\frac{\widehat \Pr (x_1, \ldots, x_t)}{\Pr (x_1, \ldots, x_t)}\right|\leq 1+\epsilon\\
\Rightarrow & 1-\epsilon \leq\left|\frac{\widehat \Pr (x_t|x_{1:t-1}) \widehat \Pr (x_{1:t-1})}{\Pr (x_t|x_{1:t-1})\Pr (x_{1:t-1})}\right|\leq 1+\epsilon\\
\Rightarrow & \frac{1-\epsilon}{1+\epsilon}\leq\left|\frac{\widehat \Pr (x_t|x_{1:t-1})}{\Pr (x_t|x_{1:t-1})}\right|\leq \frac{1+\epsilon}{1-\epsilon}
\end{align*}
and using the fact that for small enough $x$ we have $\frac{1+x}{1-x}\leq 1+3x$ and $1-3x\leq \frac{1-x}{1+x}$, plus the fact that $\epsilon_0\leq \frac{\epsilon}{6}$ we have
\begin{align*}
\Rightarrow & 1-3 \epsilon\leq\left|\frac{\widehat \Pr (x_t|x_{1:t-1})}{\Pr (x_t|x_{1:t-1})}\right| \leq 1+3 \epsilon\\
\Rightarrow & \frac{1}{1+3\epsilon}\leq\left|\frac{\Pr (x_t|x_{1:t-1})}{\widehat \Pr (x_t|x_{1:t-1})}\right|\leq \frac{1}{1-3\epsilon}\\
\end{align*}
and using a similar fact from above that for small enough $x$, $\frac{1}{1-x}\leq 1+2x$, we get
\begin{align*}
\Rightarrow & \left|\frac{\Pr (x_t|x_{1:t-1})}{\widehat \Pr (x_t|x_{1:t-1})}\right|\leq 1+ 6 \epsilon\\
\Rightarrow & \ln\left[\frac{\widehat \Pr (x_t|x_{1:t-1})}{\Pr (x_t|x_{1:t-1})}\right]\leq \ln(1+6\epsilon)\leq 6 \epsilon\\
\Rightarrow & \sum_{x_1, \ldots, x_t} \Pr(x_1, \ldots, x_t) \ln\left[\frac{\widehat \Pr (x_t|x_{1:t-1})}{\Pr (x_t|x_{1:t-1})}\right]\\
&~~~~~~~ \leq 6\epsilon\sum_{x_1, \ldots, x_t} \Pr(x_1, \ldots, x_t)\\
\Rightarrow & E \ln\left[\frac{\widehat \Pr (x_t|x_{1:t-1})}{\Pr (x_t|x_{1:t-1})}\right]\leq 6 \epsilon
\end{align*}
\hfill $\Box$

Define $ J \equiv 2m\sqrt{\frac{2\log \frac{2m}{\delta}}{N}}$ to
 simplify the following statements.  The proof proceeds in two steps.
First lemma \ref{lem:equiv} converts the sample complexity bound into
a more useful bounds
on $\Lambda$ and $\sigma_m$.  Then lemma \ref{lem:main} uses these bounds to show the
theorem. 
\begin{lemma} \label{lem:equiv} If
\begin{displaymath}
N  \geq  \frac{128 m^2}{(\sqrt[2t+3]{1 + \epsilon}-1)^2 \Lambda^2 \sigma_m^4} \log \left(\frac{2m}{\delta}\right)
\end{displaymath}
then
\begin{eqnarray}
\Lambda & \geq & \frac{3J}{\sigma_m^2(\sqrt[2t+3]{1 +
\epsilon}-1)} \label{eq:lambda:condition}\\
\sigma_m &\ge & 4J\label{eq:sigma:condition}
\end{eqnarray}
\end{lemma}
The proof is straightforward and given in the appendix.

\begin{lemma}\label{lem:main} If equation (\ref{eq:sample:complexity})
of Theorem \ref{thm:main} is replaced by (\ref{eq:lambda:condition})
and (\ref{eq:sigma:condition}) then the results of the theorem follow.
\end{lemma}

{\bf Proof of Lemma \ref{lem:main}:} Our estimator (see equation \ref{eq:phat}) can be written as
\begin{displaymath}
\widehat \Pr (x_1,\ldots, x_t)
 = \widehat \mu^\top \widehat \Sigma^{-1} \widehat K(y_t) \widehat
\Sigma^{-1}\cdots \widehat K(y_1) \widehat \Sigma^{-1} \widehat
\mu 
\end{displaymath}
We can rewrite this matrix product as
\begin{eqnarray*}
\widehat \Pr (x_1,\ldots, x_t) =& \\
                  \sum_{i_1=1}^m\cdots \sum_{i_{2t+3}=1}^m
                  &[\widehat \mu]_{i_1}
                  [\widehat \Sigma^{-1}]_{i_1,i_2} 
                  [\widehat K(y_t)]_{i_2,i_3}
                  [\widehat \Sigma^{-1}]_{i_3,i_4}\\
                  &\cdots 
                  [\widehat \mu]_{i_{2t+3}}
\end{eqnarray*}
The components $[\widehat{K}(y)]_{a,b}$ can be written as a scalar sum as:
\begin{eqnarray*}
[\widehat{K}(y)]_{a,b} = y_1 [\widehat K]_{a,b,1} + y_2 [\widehat K]_{a,b,2} + \ldots + y_m [\widehat K]_{a,b,m}
\end{eqnarray*}
So,
\begin{eqnarray*}
&\widehat \Pr (x_1,\ldots, x_t) =~~~~~~~~~~~~~~~~~~~~~~~~~~~~~~~~~~~~~~~~~~~~~~~~~ \\
                  &\sum_{\substack{i_1,\ldots,i_{2t+3}\\ j_1,\ldots,j_t}}
                  [\widehat \mu]_{i_1}
                  [\widehat \Sigma^{-1}]_{i_1,i_2} 
                  [\widehat K]_{i_2,i_3,j_1} [y_t]_{j_1}~~~~~~~ \\
                  &\cdot [\widehat \Sigma^{-1}]_{i_3,i_4} 
                  [\widehat K]_{i_5,i_6,j_2} [y_{t-1}]_{j_2} \;
                  \cdots 
                  [\widehat \mu]_{i_{2t+3}}
\end{eqnarray*}
This is just a sum of a product of scalars.  Lemma \ref{lem:element:bound}
(stated precisely and proven in the appendix) shows that accuracy of our estimates of all elements of $\mu$,
 $\Sigma^{-1}$ and $K()$ are bounded by $3J/\sigma_m^2$ with
 probability $1 - \delta$.  

Each term in the product can be rewritten as
\begin{eqnarray*}
\widehat \theta &=& \theta\left(1+\frac{\widehat \theta - \theta}{\theta}\right)
\end{eqnarray*}
and so our products can be thought of as, instead of a product of
observed quantities, the product of the theoretical quantities times
some relative error term.  We can bound this relative error term for
all entries, which will allow it to factor out nicely over all
summands, giving us a relative error term for our overall
probability. 

Again thinking of $\theta$ as a generic item in $\mu$, $\Sigma$, or
$K()$, then above has shown that $|\widehat\theta - \theta| \le 3J/\sigma^2_m$
and so the relative error of each term is bounded as
\begin{displaymath}
1-\frac{3J}{\sigma_m^2\theta} \le \; \frac{\widehat \theta}{\theta} \; \leq 1 + \frac{3J}{\sigma_m^2\theta}
\end{displaymath}
which will hold for all terms with probability $1 - \delta$.  Since $|\theta| \ge \Lambda$, we see that 
\begin{displaymath}
1-\frac{3J}{\sigma_m^2\Lambda} \le \; \frac{\widehat \theta}{\theta} \;
\leq 1 + \frac{3J}{\sigma_m^2\Lambda}
\end{displaymath}
Since our $\widehat{Pr}()$ is a product of $2t+3$ such terms, we see
that
\begin{displaymath}
\left(1-\frac{3J}{\sigma_m^2\Lambda}\right)^{2t+3} \le \frac{\widehat \Pr()}{\Pr()} \leq \left(1 + \frac{3J}{\sigma_m^2\Lambda}\right)^{2t+3}
\end{displaymath}
So by our bound on $\Lambda$, we have 
\begin{displaymath}
1- \epsilon \le  \frac{\widehat \Pr()}{\Pr()} \leq 1 + \epsilon
\end{displaymath}
holds with probability $1 - \delta$.
\hfill $\Box$

The sample complexity bound in Theorem \ref{thm:main} relies on
knowing unobserved parameters of the problem.  To avoid this, we 
modify Lemma \ref{lem:main} to make it observable. In other words, 
we convert the assumptions of sample complexity into a checkable condition.

\begin{corollary} \label{cor:main} Let $X_t$ be generated by an $m \ge 2$ state HMM.
Suppose we are given a $U$ which has the property that
$\hbox{range}(O) \subset \hbox{range}(U)$. Suppose we use
equation (\ref{eq:phat}) to estimate the probability based on $N$
independent triples.  Then with probability $1-\delta$, if the following two inequalities hold
\begin{eqnarray}
\widehat{\Lambda} \; \widehat{\sigma}_m^2\hspace{-.5em} & \geq &\hspace{-.5em}
\left(12m + \frac{6m}{(\sqrt[2t+3]{1 +
\epsilon}-1)}\right)\sqrt{\frac{2 \log \frac{2m}{\delta}}{N}} \quad \label{eq:cd:1}\\
\widehat{\sigma}_m &\ge & 10 m\sqrt{\frac{2 \log \frac{2m}{\delta}}{N}}. \label{eq:cd:2}
\end{eqnarray}
then
\begin{eqnarray*}
1-\epsilon\leq\left|\frac{\widehat \Pr (x_1, \ldots, x_t)}{\Pr (x_1, \ldots, x_t)}\right|\leq 1+\epsilon
\end{eqnarray*}
\end{corollary}

Proof:

Two technical lemma's are needed for this corollary: Lemma
\ref{lem:element:bound} and Lemma \ref{lem:estimates}.  They are
stated and proved in the supplemental material.  Lemma
\ref{lem:element:bound} basically says that with high probability,
each element of $\mu$, $\Sigma$ and $K()$ is estimated accurately.
This is then used in Lemma \ref{lem:estimates} to show that $\Lambda$
and $\sigma_m$ are estimated accurately.  

Define the event ${\cal A}$ to be the set where all the estimates
given in Lemma \ref{lem:element:bound} hold.  This event happens with
probability $1-\delta$.  On this event from Lemma \ref{lem:estimates}
we know $\sigma_m \ge \frac{4}{5} \widehat{\sigma}_m$, so $\sigma_m^2
\ge \frac{1}{2} \widehat{\sigma}_m^2$.  Hence
\begin{displaymath}
\widehat{\Lambda}  \geq  \frac{6 m}{\sigma_m^2(\sqrt[2t+3]{1 +
\epsilon}-1)}\sqrt{\frac{2 \log \frac{2m}{\delta}}{N}} +  \frac{6
m}{\sigma_m^2}\sqrt{\frac{2\log \frac{2m}{\delta}}{N}},
\end{displaymath}
thus on the set ${\cal A}$ if (\ref{eq:cd:1}) and (\ref{eq:cd:2})
hold, then we see that (\ref{eq:lambda:condition}) and
(\ref{eq:sigma:condition}) both hold and so we can apply Theorem \ref{thm:main}.
  We can now use Theorem \ref{thm:main} to generate
our claim on the accuracy of our probability bound.  Technically, this
proof as given only shows that our corollary holds with probability $1 - 2\delta$.  But
since the set where Theorem \ref{thm:main} fails is exactly ${\cal
A}^c$, the probability lower bound is $1 - \delta$.

\hfill $\Box$

The advantage of the corollary is that the left hand sides of the
two conditions are observable and the right hand sides involve known
quantities.  Hence one can tell if the condition is true or not--it
doesn't require knowing unobserved parameters. Note that the statement
is of the form $Pr(A \Rightarrow B) \ge 1 - \delta$ so interpretation
must be done carefully.

\section{Discussion: effect of $\Lambda$ and $\sigma_m$ on accuracy}

As discussed above, $\sigma_m$ and $\Lambda$ have different
effects on sample complexity.  As $\sigma_m$ approaches zero, model
estimation becomes intrinsically hard; some problems do not admit easy
estimation.  In contrast, role of $\Lambda$ in sample complexity is
more of an artifact. As $\Lambda$ approaches zero, the relative error
can be arbitrarily large, even if the estimated model is good in the
sense that the probability estimates are highly accurate.

The problem with $\Lambda$ can be addressed in a couple ways.  In this
section, we show that estimating a likelihood ratio rather than the
sequence probabilities gives improves relative accuracy bounds. An
alternate approach, which we do not pursue here, relies on the
observation that $\Lambda$ depends on the (underspecified) matrix
$\hat{U}$, and that one can thus search for a rotation and rescaling of
the matrix $\hat{U}$ that increases $\Lambda$.

\subsection{Likelihood instead of probabilities}
\label{sec:likelihood}
Obscure words correspond to rows of the observation matrix with very
small values throughout the row.  If we were interested in only
estimating the probability of such a word, then these are the easy
words--basically guess zero or close to it.  But, since we would like
to estimate the relative probability accurately, these words are the
most challenging.  Further, such small probabilities would make
computing conditional probabilities unstable since they would then
become basically ``0/0.''  Further, since the values
are all small in $O$ and in $U$, they do not significantly improve our estimates of
$\mu$, $\Sigma$ and $K()$  since they are essentially
zeros.  Both of these problems can be fixed by considering the problem
of estimating a likelihood ratio instead of a probability.  So define:
\begin{displaymath}
\lambda_q(x_1,\ldots,x_t) = \frac{Pr(x_1, x_2, \ldots, x_t)}{P_1(x_1)
P_1(x_2) \cdots P_1(x_t)}
\end{displaymath}
The $P_1(x)$ could be taken to be the marginal probability of observing $x$.
  It does not, in fact, have to be a probability--just any weighting
 which helps condition our matrix $\Sigma$ and our tensor $K()$.  We
can then use a modified version of 
 $O$ and $U$ in  all our existing lemma's and theorems.  The
 precise statement of these modified versions are in the appendix.
  What changes is that now $\Lambda$ is much larger
 and hence our relative accuracy will be greatly improved.  This fact
 is shown in the empirical section.

\subsection{Empirical estimates of $\Lambda$ and $\sigma_m$}

\begin{figure}[htbp]
\begin{center}
\hspace{-.25in}
\includegraphics[width=2.75in]{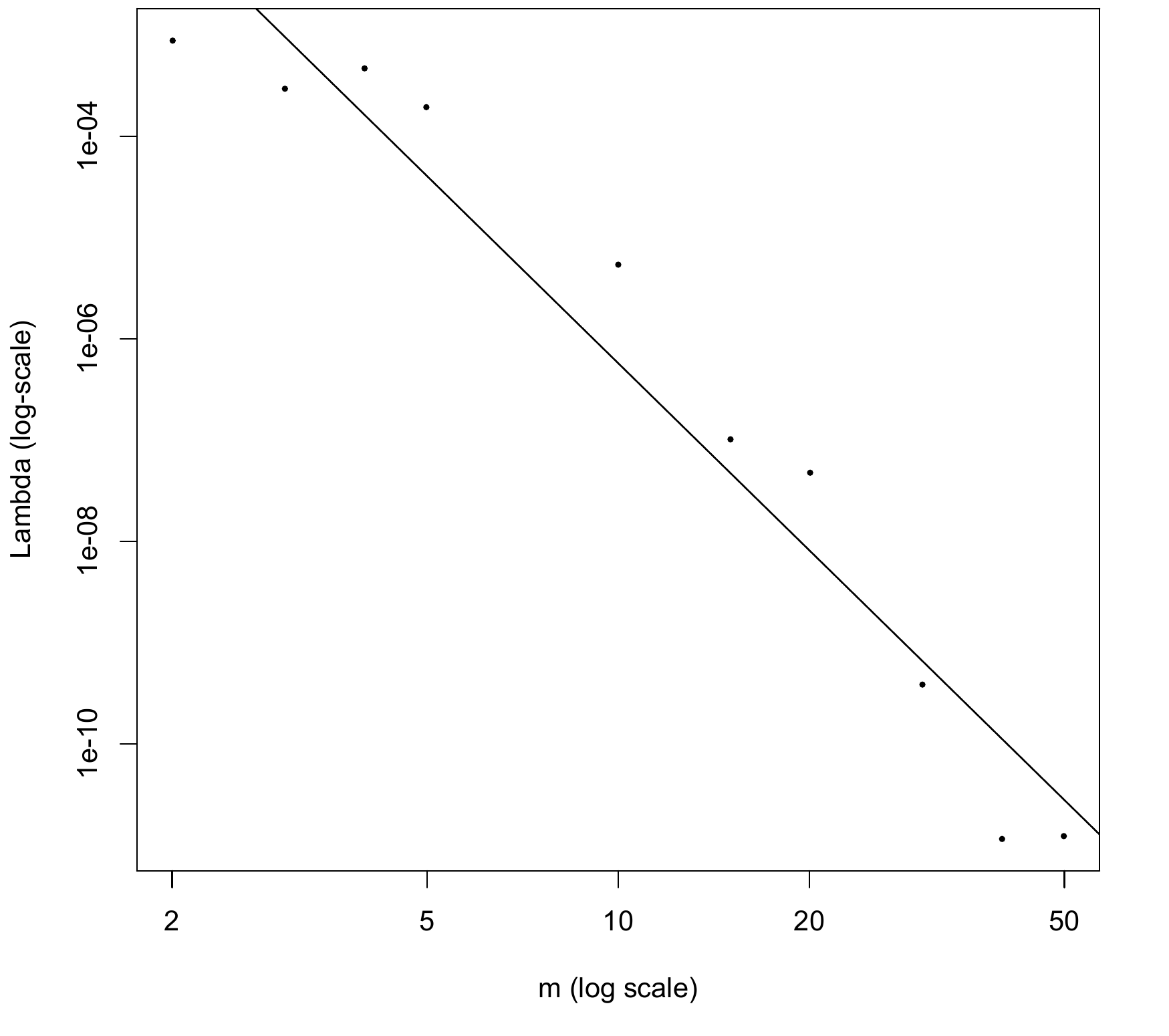}
\hspace{-.25in}
\includegraphics[width=2.75in]{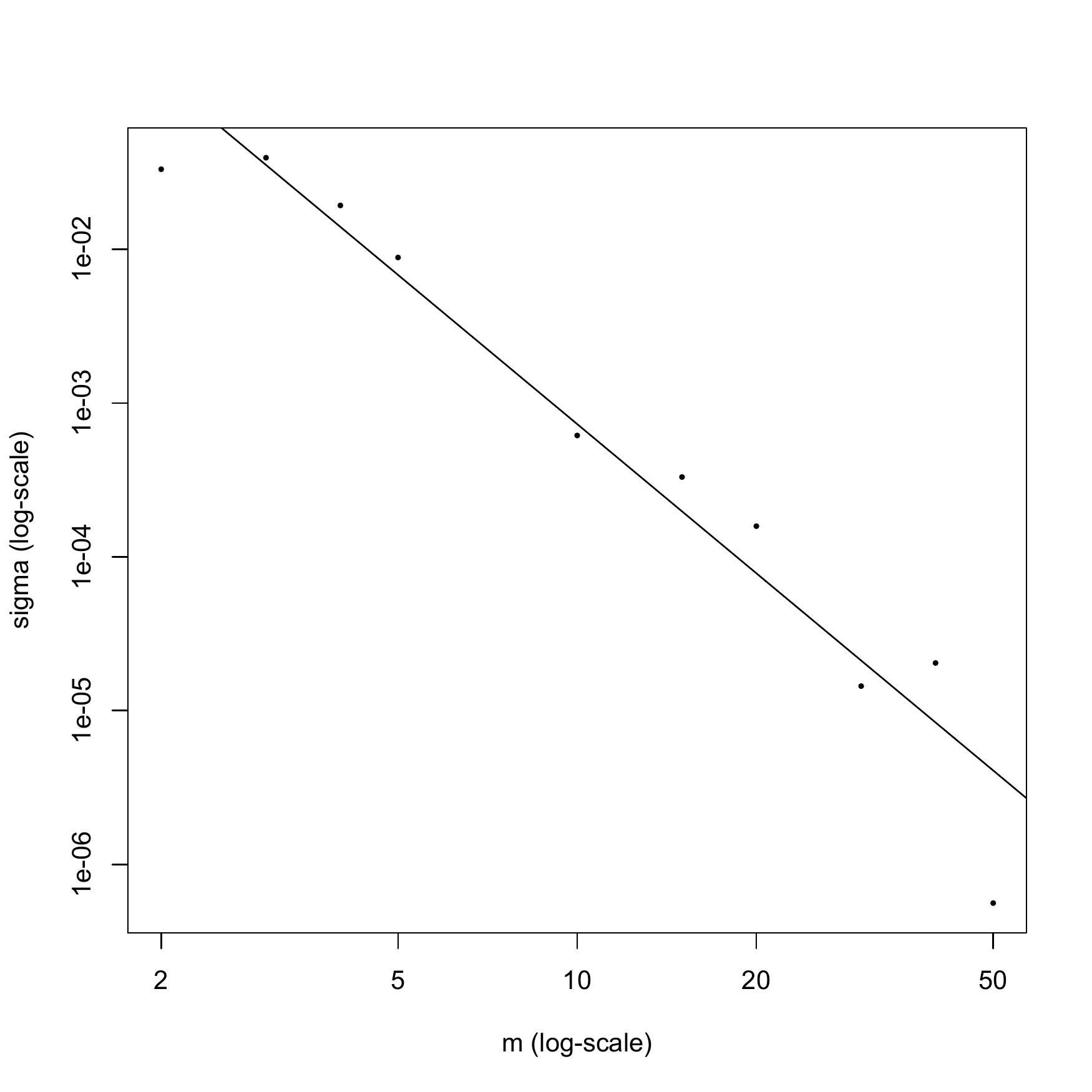}
\end{center}
\caption{{\bf First graph:} $\Lambda$ vs $m$, generated using
vocabulary size $20,000$, Slope $\approx -6$.  {\bf Second graph:} $\widehat \sigma$ vs $m$, generated
 using vocabulary size of $10,000$, Slope $\approx -3.2$}
\label{fig:empirical}
\end{figure}

Figure \ref{fig:empirical} shows estimates of $\widehat \Lambda$ and
$\widehat \sigma_m$, using the Internet as the corpus as summarized in the
Google n-gram dataset\footnote{http://googleresearch.blogspot.com/2006/08/all-our-n-gram-are-belong-to-you.html}, which contains frequencies of the most frequent 1-grams to 5-grams
occurring on the web.  Details on how the figures were generated can
be found in the supplementary material.
As the size, $m$, of the reduced dimension space is increased,
smaller and smaller singular values, $\sigma_m$, occur in the model,
and the value $\Lambda$ of the smallest parameter in the model decreases.
Empirically, both fall off with a power of $m$, giving
straight lines on the log-log plot.
This data indicates a large sample complexity, the reduction of which will be a focus of future work.

%%%%%%%%%%%%%%%%%%%%%%%%%%%%%%%%%%%%%%%%%%%%%%%%%%%%%%
\section{Prior work and conclusion}

Recently, ideas have been proposed that push spectral learning of HMMs
in several
different directions. \cite{bootshilbert} provides a kernelized
spectral algorithm that allows for learning an HMM in any domain in
which there exists a kernel.  This allows for learning
of an HMM with continuous output without the need for discretization.
\cite{boots2011online} provides an analogous algorithm that enables
online learning for Transformed Predictive State Representations, and
hence the setup in \cite{hsu}.
Finally, \cite{siddiqi2009reduced} directly extends \cite{hsu} by
relaxing the requirement that the transition matrix $T$ be of rank
$m$, but instead allows rank $k\leq m$, creating a Reduced-Rank HMM
(RR-HMM), and then applying the algorithm from \cite{hsu} to learn the
observable representation of this RR-HMM.

All of the above extensions preserve the basic structure of the
tensor $B_x$, which updates the hidden state estimate (or more
precisely, a linear transformation of it) based on the most recent
observation $x$. In this paper, we replace $B_x$ with a tensor $C(y)$,
which updates the hidden state estimate using a low dimensional projection
$y$ of the observation $x$. $C(y)$ contains only $m^3$ terms, in 
contrast to the $m^2 v$ terms contained in $B_x$. Reducing the number
of parameters to be estimated has both computational and statistical efficiency
advantages, but requires some changes to the proofs in \cite{hsu}.
While making these changes, we also give proofs that are simpler,
that only use conditions that are checkable from the data, and that bound
the relative, rather than absolute error.

This paper focused on the simplest case, in which HMMs have discrete
states and discrete observations and in which the observations are
reduced to the same sized space as the hidden state, but our approach
can be generalized in all of the ways described above.

%\section{Conclusion}

We have presented an improved spectral method for estimating HMMs. By
using a tensor $C_y$ that depends on the reduced rank $y$ instead of
the full observed $x$ in the $B_x$ tensor used by~\cite{hsu}, we
reduced the number of parameters to be estimated by a factor of the
ratio of the size of the vocabulary divided by the size of the hidden
state. This reduction has corresponding benefits in the sample
complexity.  
We also showed that the sample complexity depends critically upon
$\sigma_m$, the smallest singular value of the covariance matrix
$\Sigma$. As $\sigma_m$ becomes small, the HMM becomes increasingly
hard to identify, and increasing numbers of samples are needed.

\bibliography{bib}

\begin{thebibliography}{15}
\providecommand{\natexlab}[1]{#1}
\providecommand{\url}[1]{\texttt{#1}}
\expandafter\ifx\csname urlstyle\endcsname\relax
  \providecommand{\doi}[1]{doi: #1}\else
  \providecommand{\doi}{doi: \begingroup \urlstyle{rm}\Url}\fi

\bibitem[Baum et~al.(1970)Baum, Petrie, Soules, and
  Weiss]{baum1970maximization}
Baum, L.E., Petrie, T., Soules, G., and Weiss, N.
\newblock A maximization technique occurring in the statistical analysis of
  probabilistic functions of markov chains.
\newblock \emph{The annals of mathematical statistics}, 41\penalty0
  (1):\penalty0 164--171, 1970.

\bibitem[Boots \& Gordon(2011)Boots and Gordon]{boots2011online}
Boots, B. and Gordon, G.J.
\newblock An online spectral learning algorithm for partially observable
  nonlinear dynamical systems.
\newblock \emph{AAAI}, 2011.

\bibitem[Boots et~al.(2010)Boots, Siddiqi, Gordon, and Smola]{bootshilbert}
Boots, B., Siddiqi, S.M., Gordon, G., and Smola, A.
\newblock Hilbert space embeddings of hidden markov models.
\newblock \emph{Proc. 27th Intl. Conf. on Machine Learning (ICML)}, 2010.

\bibitem[Carlyle \& Paz(1971)Carlyle and Paz]{carlyle1971realizations}
Carlyle, J.W. and Paz, A.
\newblock Realizations by stochastic finite automata.
\newblock \emph{Journal of Computer and System Sciences}, 5\penalty0
  (1):\penalty0 26--40, 1971.

\bibitem[Dempster et~al.(1977)Dempster, Laird, and Rubin]{dempster1977maximum}
Dempster, A.P., Laird, N.M., and Rubin, D.B.
\newblock Maximum likelihood from incomplete data via the em algorithm.
\newblock \emph{Journal of the Royal Statistical Society. Series B
  (Methodological)}, 39\penalty0 (1):\penalty0 1--38, 1977.

\bibitem[Fliess(1974)]{fliess1974matrices}
Fliess, M.
\newblock Matrices de hankel.
\newblock \emph{J. Math. Pures Appl}, 53\penalty0 (197-222):\penalty0 423,
  1974.

\bibitem[Geman \& Geman(1984)Geman and Geman]{geman}
Geman, Stuart and Geman, Donald.
\newblock Stochastic relaxation, gibbs distributions, and the bayesian
  restoration of images.
\newblock \emph{Pattern Analysis and Machine Intelligence, IEEE Transactions
  on}, PAMI-6\penalty0 (6):\penalty0 721 --741, nov. 1984.
\newblock ISSN 0162-8828.
\newblock \doi{10.1109/TPAMI.1984.4767596}.

\bibitem[Hoeffding(1963)]{Hoeffding1963}
Hoeffding, Wassily.
\newblock Probability inequalities for sums of bounded random variables.
\newblock \emph{Journal of the American Statistical Association}, 58\penalty0
  (301):\penalty0 pp. 13--30, 1963.
\newblock ISSN 01621459.
\newblock URL \url{http://www.jstor.org/stable/2282952}.

\bibitem[Hsu et~al.(2009)Hsu, Kakade, and Zhang]{hsu}
Hsu, Daniel, Kakade, Sham~M., and Zhang, Tong.
\newblock A spectral algorithm for learning hidden markov models.
\newblock \emph{COLT}, 2009.

\bibitem[Jaeger(2000)]{jaeger}
Jaeger, Herbert.
\newblock Observable operator models for discrete stochastic time series.
\newblock \emph{Neural Computation}, 12(6), 2000.

\bibitem[Littman et~al.(2002)Littman, Sutton, and Singh]{littman2002predictive}
Littman, M.L., Sutton, R.S., and Singh, S.
\newblock Predictive representations of state.
\newblock \emph{Advances in neural information processing systems}, 2:\penalty0
  1555--1562, 2002.

\bibitem[Rabiner(1989)]{rabiner1989tutorial}
Rabiner, L.R.
\newblock A tutorial on hidden markov models and selected applications in
  speech recognition.
\newblock \emph{Proceedings of the IEEE}, 77\penalty0 (2):\penalty0 257--286,
  1989.

\bibitem[Schutzenbeegeb(1961)]{schutzenbeegeb1961definition}
Schutzenbeegeb, MP.
\newblock On the definition of a family of automata.
\newblock \emph{Information and control}, 4\penalty0 (2-3), 1961.

\bibitem[Siddiqi et~al.(2009)Siddiqi, Boots, and Gordon]{siddiqi2009reduced}
Siddiqi, S.M., Boots, B., and Gordon, G.J.
\newblock Reduced-rank hidden markov models.
\newblock \emph{Arxiv preprint arXiv:0910.0902}, 2009.

\bibitem[Terwijn(2002)]{terwijn2002learnability}
Terwijn, S.
\newblock On the learnability of hidden markov models.
\newblock \emph{Grammatical Inference: Algorithms and Applications}, pp.\
  344--348, 2002.

\end{thebibliography}
\bibliographystyle{icml2012}

\newpage

\section*{APPENDIX- SUPPLEMENTAL MATERIAL}

\begin{lemmaNoNum}[{\bf Restatement of Lemma \ref{lem:representation}}] Assume the hidden state is of dimension $m$ and the rank
of $O$ is also $m$. Then:
\begin{equation*}
Pr(x_1 ,x_2, \ldots, x_t)
      = 1^\top A_{x_t} A_{x_{t-1}} \cdots A_{x_1} \pi \tag{\ref{eqn:orig}}
\end{equation*}
\begin{equation*}
Pr(x_1 ,x_2, \ldots, x_t)
      = b_\infty^\top B_{x_t} B_{x_{t-1}} \cdots B_{x_1} b_1
\tag{\ref{eqn:hsu:lemma}} 
\end{equation*}
\begin{equation*}
Pr(x_1 ,x_2, \ldots, x_t)
      = c_\infty^\top C_{y_t} C_{y_{t-1}} \cdots C_{y_1} c_1 \tag{\ref{eqn:new}}
\end{equation*}
Where (\ref{eqn:hsu:lemma}) requires $U^\top O$ to be invertible, and
(\ref{eqn:new}) requires $\hbox{range}(O) \subset \hbox{range}(U)$.
\end{lemmaNoNum}

{\bf Proof: }

As pointed out in the main text, \cite{jaeger} showed (\ref{eqn:orig}),
and  \cite{hsu} showed (\ref{eqn:hsu:lemma}).  To show (\ref{eqn:new}),
we will first write the characteristics $\mu$, $\Sigma$ and $K$ in
terms of the theoretical matrices, $T$, $O$, $U$, and $\pi$:
\begin{eqnarray*}
\mu &=& U^\top O\pi \\
\Sigma &=& U^\top O\; T\; \diag(\pi) \; O^\top U \\
\Sigma^{-1} &=& (O^\top U)^{-1} \; \diag(\pi)^{-1} \; T^{-1} \; (U^\top O)^{-1} \\
K(y) &=& U^\top O \; T\; \diag(O^\top U y) \; T\; \diag(\pi) \; O^\top  U
\end{eqnarray*}
By definition, we have \\
\begin{displaymath}
c_1 \equiv \mu = U^\top  O \; \pi
\end{displaymath}
likewise,
\begin{eqnarray*}
c^\top _\infty & \equiv & \mu^\top  \Sigma^{-1} \\
&=&\left(\pi^\top O^\top U\right) \left( (O^\top U)^{-1} \; \diag(\pi)^{-1}\right. \\
&&~~~~~~~~~~\left.\cdot \; T^{-1} \; (U^\top O)^{-1} \right)\\
&=&\pi^\top  \; \diag(\pi)^{-1} \; T^{-1} \; (U^\top O)^{-1} \\
&=&{\bf 1}^\top  T^{-1} (U^\top  O)^{-1}\\
&=&{\bf 1}^\top   (U^\top  O)^{-1}
\end{eqnarray*}
For $C$,
\begin{eqnarray*}
C(y) & = &  K(y) \; \Sigma^{-1} \\
&=& U^\top   O \; T \; \diag(O^\top Uy) \; \\
&&~~~~~~~~~~~\cdot \; T \;
\diag(\pi) \; O^\top U \; \Sigma^{-1} \\
&=& U^\top   O \; T \; \diag(O^\top Uy) (U^\top O)^{-1} 
\end{eqnarray*}
Note that $UU^\top$ is a projection operator and since its range is the
 same as that of $O$ we have $O^\top UU^\top = O^\top$.
So, if $y = U^\top \delta_x$, then:
\begin{eqnarray*}
C(y) & = &  U^\top   O \; T \; \diag(O^\top U U^\top \delta_x) (U^\top O)^{-1} \\
    & = &  U^\top   O \; T \; \diag(O^\top \delta_x) (U^\top O)^{-1} \\
    & = &  U^\top   O \; A_x \; (U^\top O)^{-1} 
\end{eqnarray*}

Thus (\ref{eqn:new}) follows from a telescoping product.  

\hfill $\Box$

{\bf Proof of lemma \ref{lem:equiv}:}

The proof is simply algebraic manipulation.  We have 
\begin{displaymath}
N \geq  \frac{128 m^2}{(\sqrt[2t+3]{1 + \epsilon}-1)^2 \Lambda^2 \sigma_m^4} \log \left(\frac{2m}{\delta}\right)
\end{displaymath}
which implies that
\begin{eqnarray*}
\Lambda^2 & \geq & \frac{128 m^2}{(\sqrt[2t+3]{1 + \epsilon}-1)^2 N \sigma_m^4} \log \left(\frac{2m}{\delta}\right)\\
& \geq & \frac{72 m^2}{(\sqrt[2t+3]{1 + \epsilon}-1)^2 N \sigma_m^4} \log \left(\frac{2m}{\delta}\right)
\end{eqnarray*}
and taking the square root and making the relevant substitution for J we have
\begin{eqnarray*}
\Lambda \geq \frac{3J}{\sigma_m^2(\sqrt[2t+3]{1 +
\epsilon}-1)}\\
\end{eqnarray*}
To show the bound for $\sigma_m$ we have that
\begin{displaymath}
N \geq  \frac{128 m^2}{(\sqrt[2t+3]{1 + \epsilon}-1)^2 \Lambda^2 \sigma_m^4} \log \left(\frac{2m}{\delta}\right)
\end{displaymath}
and noting that $\Lambda < 1$ and $\sqrt[2t+3]{1 + \epsilon}-1 < 1$, 
\begin{eqnarray*}
\sigma^4 \geq \frac{128 m^2}{N} \log \left(\frac{2m}{\delta}\right)
\end{eqnarray*}
Taking the square root of both sides and making the relevant substitution, we get
\begin{eqnarray*}
\sigma_m^2 \geq 4J
\end{eqnarray*}
and since $\sigma_m < 1$ implies $\sigma_m^2 < \sigma_m$ then we get the desired inequality.
\hfill $\Box$

\begin{lemma}\label{lem:element:bound}
Our estimates of all elements of $\mu$, $\Sigma^{-1}$ and $K()$ are bounded
by $3J/\sigma_m^2$ with probability $1 - \delta$, where $ J \equiv 2m\sqrt{\frac{2\log \frac{2m}{\delta}}{N}}$.
\end{lemma}

{\bf Proof:}

We first derive absolute bounds for each entry of $\mu$, $\Sigma$
 and $K()$.  To handle all three of them at the same time, we will
 generically call any one of these three ``$\theta$'' and its estimate
 $\widehat{\theta}$.  Suppose that $\widehat{\theta}$ has $g$ entries that
 are taking the mean with $N$ observations all of which are bounded
 between $-1$ and $1$.  Then, for each entry we have from \cite{Hoeffding1963} that 
\begin{eqnarray*}
Pr(|\widehat{\theta}_i-\theta_i| > \epsilon)\leq 2e^{\frac{-N\epsilon^2}{2}}
\end{eqnarray*}
and so
\begin{eqnarray*}
Pr(\exists ~ i ~ \mathrm{s.t.} ~ |\widehat{\theta}_i - \theta_i| > \epsilon)\leq 2ge^{\frac{-N\epsilon^2}{2}}
\end{eqnarray*}
and setting $2ge^{\frac{-N\epsilon^2}{2}}=\delta$ we solve that
$\epsilon=\sqrt{\frac{2\log \frac{2g}{\delta}}{N}}$ so with probability
$1-\delta$ we have that  
\begin{eqnarray*}
\forall i \quad |\widehat{\theta}_i-\theta_i| \leq\sqrt{\frac{2\log \frac{2g}{\delta}}{N}}.
\end{eqnarray*}
Note that for $\mu$, $\Sigma$ and $K()$ we have a vector, a matrix and a tensor that are
estimated as $E(Y_1)$, $E(Y_1Y_2^\top)$ and $E(Y_3 Y_1^\top Y_2^\top)$
respectively with $m$, $m^2$ and
$m^3$ entries respectively, we see that the total number of entries in
all three of them is less than $m^4$.  (Except in the trivial case
where $m = 1$.  But this corresponds to the data being IID and so
doesn't count as a HMM.)  So all three of the following hold simultaneously
with probability $1 - \delta$:
\begin{eqnarray}
\forall i     \quad |\widehat \mu_i - \mu_i| & \leq & \sqrt{\frac{8\log \frac{2m}{\delta}}{N}} \nonumber \\
\forall i,j   \quad |\widehat \Sigma_{ij} -
\Sigma_{ij}|&\leq&\sqrt{\frac{8 \log \frac{2m}{\delta}}{N}}\label{eq:sigma:bound}\\
\forall i,j,j \quad |[\widehat K]_{ijk} - [K]_{ijk}|
&\leq&\sqrt{\frac{8 \log \frac{2m}{\delta}}{N}} \nonumber
\end{eqnarray}
Lastly we need to bound $\Sigma^{-1}$.  We will start by bounding the
norm of $\widehat \Sigma - \Sigma$.  By (\ref{eq:sigma:bound})  we see $||\widehat
\Sigma - \Sigma||_{\max} \le \sqrt{\frac{8 \log
\frac{2m}{\delta}}{N}}$, by the relationship $||M||_2\leq m
||M||_{\max}$ for $m \times m$ square matrices, we get the desired result.

From this bound on $||\widehat \Sigma - \Sigma||_2$ and lemma 20 of \cite{hsu} we have that
\begin{equation}
\label{eq:sigma:m:bound}
|\widehat \sigma_m - \sigma_m|\leq J
\end{equation}
where $\sigma_m$ is the smallest singular value for $\Sigma$.  By their Lemma 23 we then have that
\begin{eqnarray*}
||\widehat \Sigma^{-1}-\Sigma^{-1}||_2\leq \frac{1+\sqrt{5}}{2}\left(\frac{1}{\widehat \sigma_m - J}\right)^2 J
\end{eqnarray*}
By assumption $\sigma_m > 4 J$, we see $\sigma_m - J > 3\sigma_m/4$.
Thus from the algebra that $\frac{1+\sqrt{5}}{2}(\frac{4}{3})^2 \le 3$,
we see
\begin{eqnarray*}
||\widehat \Sigma^{-1}-\Sigma^{-1}||_2\leq 3J / \sigma_m^2.
\end{eqnarray*}
From $||\widehat \Sigma^{-1}-\Sigma^{-1}||_{\max} \le ||\widehat
\Sigma^{-1}-\Sigma^{-1}||_2$ we get our element-wise norm on the
errors. Since $\sigma_m \le 1$, we see that
\begin{displaymath}
3J / \sigma_m^2 \ge 3J =  3m\sqrt{\frac{8 \log
\frac{2m}{\delta}}{N}}\ge \sqrt{\frac{8\log \frac{2m}{\delta}}{N}}
\end{displaymath}

\hfill $\Box$

\begin{lemma} \label{lem:estimates} The estimates of $\Lambda$ and $\sigma_m$ have the
following accuracy:
\begin{eqnarray}
|\widehat{\Lambda} - \Lambda| & \le & \frac{6 m}{\sigma_m^2}\sqrt{\frac{2\log \frac{2m}{\delta}}{N}}
 \nonumber \label{eq:lambda:accuracy}\\
|\widehat{\sigma}_m - \sigma_m| &\le &  2m\sqrt{\frac{2 \log \frac{2m}{\delta}}{N}}. 
\nonumber \label{eq:sigma:accuracy}
\end{eqnarray}
with probability greater than $1 - \delta$.
\end{lemma}
Proof:
$\widehat{\Lambda}$ is the empirical minimum of all the 
\begin{displaymath}
\widehat{\Lambda} \equiv \min\{\min_i |\widehat{\mu}_i|,\min_{i,j} |\widehat{\Sigma}_{ij}^{-1}|,
\min_{i,j,k} |\widehat{K}_{i,j,k}|\}
\end{displaymath}
From lemma \ref{lem:element:bound} we have bounded the accuracy of the
estimate of each element of $\mu$, $\Sigma$ and $K()$, the minimum of
these will be estimated within the same accuracy.  This established
(\ref{eq:lambda:accuracy}). 

The second inequality (\ref{eq:sigma:accuracy}) was also established
in the proof of the theorem in equation (\ref{eq:sigma:m:bound}).

\hfill $\Box$

\section{Likelihood ratio version of theorem \ref{thm:main}}

In \ref{sec:likelihood} we considered the likelihood ratio as a way of
getting a better estimator.  There we used a weighting vector $p_i$
which normalized our probability.  In other words,
\begin{displaymath}
\frac{Pr(x_1, x_2, \ldots, x_t)}{p_{x_1} p_{x_2} \cdots p_{x_t}}
\end{displaymath}
It will be a bit more mathematically convenient if we instead use $q_i =
 1/\sqrt{p_i}$ instead.  So, define:
\begin{displaymath}
Q(x_{1:t}) = Q(x_1, x_2, \ldots, x_t) =q(x_1) q(x_2) \cdots q(x_t)
\end{displaymath}
Then our ``likelihood ratio'' is 
\begin{displaymath}
\lambda(x_1 ,x_2, \ldots, x_t) = Pr(x_1, x_2, \ldots, x_t) Q(x_1, x_2,
\ldots, x_t)^2
\end{displaymath}

We will think of these $q_i$'s as a vector and define 
\begin{displaymath}
O^* \equiv \diag(q) O
\end{displaymath}
and
\begin{displaymath}
A^*_x  \equiv T \diag(O^{*T} \diag(q) \delta_x)
\end{displaymath}
We will then be able to show a similar product rule as (\ref{eq:jaeger}):
\begin{displaymath}
Pr(x_{1:t}) Q^2(x_{1:t}) = 1^\top A^*_{x_t} A^*_{x_{t-1}} \cdots A^*_{x_1} \pi.
\end{displaymath}
The version of this product rule we will estimate is also similar.  We
will define $U^* = \diag(q) U$ and $y^*_t = U^{*\top} \diag(q) \delta_{x_t} = U^\top \diag(q)^2
 \delta_{x_t}$.  Our statistics are then:
\begin{eqnarray*}
\mu^*& \equiv& E(y^*_1)\\
\Sigma^* & \equiv& E(y^*_2 y^{*\top}_1)\\
K^*(a) &\equiv& E(y^*_3 y^{*\top}_1 y^{*\top}_2) a
\end{eqnarray*}
Defining our characteristics as before:
\begin{eqnarray*}
c^*_1 &\equiv & \mu^* \\
c^{*\top} _\infty & \equiv & \mu^{*\top}  \Sigma^{*-1} \\
C^*(y^*) & = &  K^*(y^*) \; \Sigma^{*-1}
\end{eqnarray*}
These can also be used to estimate $\lambda$ as the following lemma
shows: 

\begin{lemma} \label{lem:representation2} Assume the hidden state is of
 dimension $m$ and the rank of $O$ is also $m$. Then:
\begin{eqnarray}
\lambda(x_1 , \ldots, x_t) & \equiv & Pr(x_{1:t}) Q^2(x_{1:t})\nonumber\\
 &=& 1^\top A^*_{x_t} A^*_{x_{t-1}} \cdots
A^*_{x_1} \pi\nonumber\\
 &=& c^{*\top}_\infty C^*(y^*_t)\cdots C^*(y^*_1) c^*_1\label{eqn:star}
\end{eqnarray}  
Where the last equation requires\\
$\hbox{range}(O) \subset
\hbox{range}(U \diag(q))$.
\end{lemma}

Proof:

\begin{eqnarray*}
A^*_x &\equiv& T \; \diag(O^{*\top} \diag(q) \delta_x) \\
 &= & T \; \diag((\diag(q) O)^{\top} \diag(q) \delta_x) \\
 &= & T \; \diag( O^{\top} \diag(q)^2 \delta_x) \\
 &= & T \; \diag( O^{\top} \diag(q)^2 \diag(\delta_x){\bf 1}) \\
 &= & T \; \diag( O^{\top} \diag(\delta_x)^2 \diag(q)^2 {\bf 1}) \\
 &= & T \; \diag( O^{\top} \diag(\delta_x)(q^2_x)) \\
 &= & T \; \diag( O^{\top}\delta_x) \; q^2_x\\
 &= & A_x \; q^2_x\\
\end{eqnarray*}
where we have used $a^\top \diag(\delta_x) b = (a^\top \delta_x) \; (
b^\top \delta_x)$.

Our ``starred'' versions can be written in terms of the basic items
 $T$, $O$, $U$, $\pi$ and $q$:
\begin{eqnarray*}
\mu^* &=& U^\top \diag(q)^2 O\pi \\
\Sigma^* &=& U^\top  \diag(q)^2 O\; T\; \diag(\pi) \; O^\top
\diag(q)^2 U \\
\Sigma^{*-1} &=& (O^\top \diag(q)^2 U)^{-1} \; \diag(\pi)^{-1} \;\\
&&~~~~~~~~\cdot \; T^{-1} \; (U^\top  \diag(q)^2 O)^{-1} \\
K^*(x) &=& U^\top \diag(q)^2 O \; T\; \diag(O^\top \diag(q)^2 U x) \; \\
&&~~~~~~~~\cdot \; T\; \diag(\pi) \; O^\top  \diag(q)^2 U
\end{eqnarray*}
So, we have \\
\begin{displaymath}
c^*_1 \equiv \mu^* = U^\top  \diag(q)^2 O \; \pi
\end{displaymath}
likewise,
\begin{eqnarray*}
c^{*\top} _\infty & \equiv & \mu^{*\top}  \Sigma^{*-1} \\
&=&\left(\pi^\top O^\top \diag(q)^2 U\right)\\
&&~~~~~~~~~~~~\cdot \; \left( (O^\top
\diag(q)^2 U)^{-1} \; \diag(\pi)^{-1} \; \right.\\
&&~~~~~~~~~~~~~~~~~~~~~~\cdot \; \left. T^{-1} \; (U^\top  \diag(q)^2O)^{-1} \right)\\
&=&\pi^\top  \; \diag(\pi)^{-1} \; T^{-1} \; (U^\top  \diag(q)^2O)^{-1} \\
&=&{\bf 1}^\top  T^{-1} (U^\top   \diag(q)^2O)^{-1}\\
&=&{\bf 1}^\top   (U^\top   \diag(q)^2 O)^{-1}
\end{eqnarray*}
For $C^*$ we 
\begin{eqnarray*}
C^*(y) & = &  K^*(y) \; \Sigma^{*-1} \\
&=& U^\top \diag(q)^2   O \; T \; \diag(O^\top \diag(q)^2 Uy) \; \\
&& ~~~~~\cdot \; T \; \diag(\pi) \; O^\top  \diag(q)^2 U \; \Sigma^{*-1} \\
&=& U^\top \diag(q)^2   O \; T \; \diag(O^\top \diag(q)^2 Uy)\\
&&~~~~~~~~~ \cdot \; (U^\top  \diag(q)^2O)^{-1} 
\end{eqnarray*}
Note that $U^* U^{*\top}$ is an $n \times n$ projection operator.  Since its range is the
 same as that of $O^*$ we have $O^{*\top} U^*  U^{*\top} = O^{*\top}$.
So, if $y^* = U^{*\top}  \diag(q) \delta_x$, then:
\begin{eqnarray*}
C^*({y^*}) & = &  U^\top \diag(q)^2   O \; T \;\\
&&\cdot \; \diag(O^{*\top}  U^* U^{*\top}  \diag(q) \delta_x)\\
&&~~~~~~~~~~~~~\cdot \; (U^\top
\diag(q)^2 O)^{-1} \\
    & = &  U^\top \diag(q)^2   O \; T \; \diag(O^\top
\diag(q)^2\delta_x)\\
&&~~~~~\cdot \; (U^\top  \diag(q)^2 O)^{-1} \\
    & = &  (U^\top    \diag(q)^2 O) \; A^*_x \; (U^\top  \diag(q)^2 O)^{-1} 
\end{eqnarray*}

Hence equation (\ref{eqn:star}) follows by a telescoping product.

\hfill $\Box$

\begin{theorem} \label{thm:lambda} Let $X_t$ be generated by an $m \ge 2$ state HMM.
Suppose we are given a $U$ which has the property that
$\hbox{range}( O) \subset \hbox{range}(U)$ and $|U_{ij}| \le 1$. Suppose we use
equation (\ref{eqn:star}) to estimate $\lambda(x_1, x_2, \ldots, x_t)$ based on $N$
independent triples and for appropriate choice of $U^*$.  Then the following two inequalities
\begin{eqnarray}
\Lambda^* & \geq & \frac{6 m}{\sigma_m^{*2}(\sqrt[2T+3]{1 +
\epsilon}-1)}\sqrt{\frac{2\log \frac{2m}{\delta}}{N}} \label{eq:lambda:condition2}\\
\sigma_m^* &\ge & 8 m\sqrt{\frac{2 \log \frac{2m}{\delta}}{N}}.\label{eq:sigma:condition2}
\end{eqnarray}
(where $\sigma_m^*$ is the smallest eigenvalue of $\Sigma^*$) 
imply
\begin{eqnarray*}
1-\epsilon\leq\left|\frac{\widehat \lambda (x_1, \ldots, x_t)}{\lambda (x_1, \ldots, x_t)}\right|\leq 1+\epsilon
\end{eqnarray*}
or equivalently
\begin{eqnarray*}
1-\epsilon\leq\left|\frac{\widehat \Pr (x_1, \ldots, x_t)}{\Pr (x_1, \ldots, x_t)}\right|\leq 1+\epsilon
\end{eqnarray*}
holds with probability at least $1 - \delta$.

\end{theorem}

{\bf Proof:}

The proof of this goes is identical to that given for theorem
 \ref{thm:main}.  The only worry is that we have defined $y^*$'s
differently.  But since we only required $|y|\le 1$, and we have
constructed $|y^*|\le 1$, the Hoeffding inequality with elements
 of $U$ still hold for $U^*$.

\hfill $\Box$

\subsection*{Details of generating the graphs}

In lemma \ref{lem:representation2} and theorem \ref{thm:lambda} we see
that we can increase our chances of obtaining a large enough $\Lambda$
by multiplying each row of $U$ by some function of that row.  As long
as we ensure that the elements of our new $U^*$ are less than one,
then we can make a claim on the accuracy of the relative "likelihood",
and hence the relative probability, generated by our sample. 

Our figures utilize this gain in the size of $\Lambda$.  For our
corpus we use the Internet as captured by the Google n-gram dataset.
We first create a dictionary of the $v-1$ most popular tokens, as well
as an "out of vocabulary" token, for a final dictionary of size $v$.
We take $U$ to be the $U$ matrix generated by the 'thin' SVD of the
$P_{21}$ matrix generated using this vocabulary and Google 2-grams.  

From this $U$ we consider the first $m$ columns.  As per above, we can
 increase our chances of obtaining a large enough $\Lambda$ by
 maximizing the size of the entries in this new $v \times m$
 dimensional $U$ matrix, hence we multiply each row by
 $1/\max_j(|U_{i,j}|)$, ensuring that at least one of the elements in
 our matrix is exactly $1$ or $-1$.  Now, using this new matrix $U^*$
 we use the frequencies from Google 1-grams, 2-grams, and 3-grams to
 compute $\mu^*$, $\Sigma^*$, and $K^*$ respectively, where each of the $v$
 vocabulary words (including one out-of-vocabulary token) correspond
 to a row of $U^*$.  From this, we take $\Sigma^{*-1}$ and compute the
 minimum element across $\mu^*$, $\Sigma^{*-1}$ and $K^*$.

We obtain $\sigma^*_m$ in a similar way, first computing $\Sigma^*$ from
the appropriate $v \times m$ dimensional $U^*$ matrix, then taking the
SVD, recording the smallest singular value. 
\end{document}